\documentclass{article}

\usepackage{microtype}
\usepackage{graphicx}
\usepackage{subfigure}
\usepackage{booktabs} 
\usepackage{epstopdf}
\usepackage{multirow}
\usepackage{hyperref}
\usepackage{booktabs}       
\usepackage{amsfonts}       
\usepackage{nicefrac}       

\usepackage{amsmath}

\usepackage[accepted]{icml2019}

\icmltitlerunning{Interpreting Adversarially Trained Convolutional Neural Networks}

\begin{document}

\twocolumn[
\icmltitle{Interpreting Adversarially Trained Convolutional Neural Networks}




\begin{icmlauthorlist}
\icmlauthor{Tianyuan Zhang}{eecs}
\icmlauthor{Zhanxing Zhu}{math,cds,bibdr}
\end{icmlauthorlist}

\icmlaffiliation{eecs}{School of EECS, Peking University, China}
\icmlaffiliation{math}{School of Mathematical Sciences, Peking University, China}
\icmlaffiliation{cds}{Center for Data Science, Peking University}
\icmlaffiliation{bibdr}{Beijing Institute of Big Data Research}

\icmlcorrespondingauthor{Zhanxing Zhu}{zhanxing.zhu@pku.edu.cn}

\icmlkeywords{Machine Learning, ICML}

\vskip 0.3in
]



\printAffiliationsAndNotice{}  

\begin{abstract}
We attempt to interpret how adversarially trained convolutional neural networks (AT-CNNs) recognize objects. We  design systematic approaches to interpret AT-CNNs in both qualitative and quantitative ways and compare them with normally trained models. Surprisingly, we find that adversarial training alleviates the texture bias of standard CNNs when trained on object recognition tasks, and helps CNNs learn a more shape-biased representation. 
We validate our hypothesis from two aspects.  First, we compare the salience maps of AT-CNNs and standard CNNs on clean images and images under different transformations. The comparison could visually show that the prediction of the two types of CNNs is sensitive to dramatically different types of features. Second, to achieve quantitative verification, we construct additional test datasets that destroy either textures or shapes, such as style-transferred version of clean data,  saturated images and patch-shuffled ones, and then evaluate the classification accuracy of AT-CNNs and normal CNNs on these datasets.  
Our findings shed some light on why AT-CNNs are more robust than those normally trained ones and contribute to a better understanding of adversarial training over CNNs from an interpretation perspective. 
\end{abstract}

\section{Introduction}
Convolutional neural networks (CNNs) have achieved great success in a variety of visual recognition tasks~\cite{krizhevsky2012imagenet, girshick2014rich, long2015fully} with their stacked local connections. A crucial issue is to understand what is being learned after training over thousands or even millions of images. This involves interpreting CNNs. 

Along this line,  some recent works showed that standard CNNs trained on ImageNet make their predictions rely on the local textures rather than long-range dependencies encoded in the shape of objects~\cite{geirhos2018imagenet,brendel2018approximating,ballester2016performance}.  
Consequently, this texture bias prevents the trained CNNs from generalizing well on those images with distorted textures but maintained shape information. \citet{geirhos2018imagenet} also showed that using a combination of Stylized-ImageNet and ImageNet can alleviate the texture bias of standard CNNs.
It naturally raises an intriguing question:

\emph{Are there any other trained CNNs are more biased towards shapes?}

Recently,  normally trained neural networks were found to be easily fooled by maliciously perturbed examples, i.e., adversarial examples~\cite{goodfellow2014explaining, kurakin2016adversarial}.  To defense the adversarial examples, adversarial training was proposed; that is, instead of minimizing the loss function over the clean example, it minimizes almost worst-case loss over the slightly perturbed examples~\cite{madry2017towards}. We name these adversarially trained networks as AT-CNNs. They were extensively shown to be able to enhance the robustness, i.e., improving the classification accuracy over the adversarial examples. Then,

\emph{What is learned by adversarially trained CNNs to make it more robust?}

In this work, in order to explore the answer to the above questions, we systematically design various experiments to interpret the AT-CNNs and compare them with normally trained models. 
We find that AT-CNNs are better at capturing long-range correlations such as shapes, and less biased towards textures than normally trained CNNs in popular object recognition datasets. This finding partially explains why AT-CNNs tends to be more robust than standard CNNs.

We validate our hypothesis from two aspects.  First, we compare the salience maps of AT-CNNs and standard CNNs on clean images and those under different transformations. The comparison could visually show that the predictions of the two CNNs are sensitive to dramatically different types of features. Second, we construct additional test datasets that destroy either textures or shapes, such as the style-transferred version of clean data, saturated images and patch-shuffled images, then evaluate the classification accuracy of AT-CNN and normal CNNs on these datasets. These sophisticated designed experiments provide a quantitative comparison between the two CNNs and demonstrate their biases when making predictions. 

To the best of our knowledge, we are the first to implement  systematic investigation on interpreting the adversarially trained CNNs, \emph{both visually and quantitatively}. Our findings shed some light on why AT-CNNs are more robust than those normally trained ones and also contribute to better understanding adversarial training over CNNs from an interpretation perspective.\footnote{Our codes are available at \href{https://github.com/PKUAI26/AT-CNN}{https://github.com/PKUAI26/AT-CNN}} 

The remaining of the paper is structured as follows. We introduce background knowledge on adversarial training and salience methods in Section~\ref{sec:pre}. The methods for interpreting AT-CNNS are described in Section~\ref{sec:methods}. Then we present the experimental results to support our findings in Section~\ref{sec:exp}. The related works and discussions are presented in Section~\ref{sec:discussion}. Section~\ref{sec:con} concludes the paper.  

\section{Preliminary}
\label{sec:pre}
\subsection{Adversarial training} 
This training method was first proposed by  \cite{goodfellow2014explaining}, which is the most successful approach for building robust models so far for defending adversarial examples~\cite{madry2017towards, sinha2017certifiable, athalye2018obfuscated, zhang2019theoretically, zhang2019you}. It can be formulated as solving a robust optimization problem~\cite{shaham2015understanding}
\begin{equation}
\min _ { \theta } \mathbb { E }_{{ ( x , y ) \sim  { \mathcal { D } } }}\left[ \max _ { \delta \in S } \ell ( f(x + \delta; \theta), y) \right],
\end{equation}
where $f(x;\theta)$ represents the neural network parameterized by weights $\theta$; the input-output pair $(x,y)$ is sample from the training set $ \mathcal { D }$; $\delta$ denotes the adversarial perturbation and $\ell(\cdot,\cdot)$ is the chosen loss function, e.g. cross entropy loss.  $S$ denotes a certain norm constraints, such as $\ell_{\infty}$ or $\ell_2$.

The inner maximization is approximated by adversarial examples generated by various attack methods. Training against a projected gradient descent (PGD,~\citet{madry2017towards}) adversary leads to state-of-the-art white-box robustness. We use PGD based adversarial training with bounded  $l_{\infty}$ and $l_2$ norm constraints. We also investigate FGSM~\cite{goodfellow2014explaining} based adversarial training.

\subsection{Salience maps} 
Given a trained neural network, visualizing the salience maps aims at assigning a \textit{sensitivity} value, sometimes also called ``attribution'', to show the sensitivity of the output  to each pixel of an input image. Salience methods can mainly be divided into \cite{ancona2018towards} \textit{perturbation-based methods}~\cite{zeiler2014visualizing, zintgraf2017visualizing} and \textit{gradient-based} method~\cite{erhan2009visualizing, simonyan2013deep, shrikumar2017learning, sundararajan2017axiomatic, selvaraju2017grad, zhou2016learning,  smilkov2017smoothgrad, bach2015pixel}. Recently \cite{adebayo2018sanity} carries out a systematic test for many of the gradient-based salience methods, and only variants of Grad and GradCAM~\cite{selvaraju2017grad} pass the proposed sanity checks. We thus choose Grad and its smoothed version SmoothGrad~\cite{smilkov2017smoothgrad} for visualization. 

Formally, let $x \in \mathbb{R}^d $ denote the input image, a trained network is a function $f : \mathbb{R}^d \to \mathbb{R}^K$, where $K$ is the total number of classes. Let $S_c$ denotes the class activation function for each class $c$.  We seek to obtain a salience map $E \in \mathbb{R}^d$. 
The \textbf{Grad} explanation  is the gradient of class activation with respect to the input image $x$,  
\begin{equation}
    E = \frac{\partial S_c(x)}{\partial x}.
\end{equation}
\textbf{SmoothGrad}~\cite{smilkov2017smoothgrad} was proposed to alleviate noises in gradient explanation by averaging over the gradient of noisy copies of an input. Thus for an input $x$, the smoothed variant of Grad, SmoothGrad can be written as 
\begin{equation}
    E = \frac{1}{n} \sum_{i=1}^n \frac{\partial S_c(x_i )}{\partial x_i}, 
    \label{eq:smoothgrad}
\end{equation}
where $x_i = x + g_i$, and $g_i$ are noise vectors drawn i.i.d from a Gaussian distribution $\mathcal{N}(0, \sigma^2) $. In all our experiments, we set $n = 100$, and the noise level ,  $\sigma/(x_{max} - x_{min}) = 0.1$. We choose  $S_c(x) = \log p_c(x)$, where $p_c(x)$ is the probability of class $c$ assigned by a classifier to input $x$.

\section{Methods}
\label{sec:methods}
In this section, we elaborate our method for interpreting the adversarially trained CNNs and comparing them with normally trained ones. Three image datasets are considered, including Tiny ImageNet\footnote{\url{https://tiny- imagenet.herokuapp.com/}}, Caltech-256~\cite{griffin2007caltech} and CIFAR-10.  

We first visualize the salience maps of AT-CNNs and normal CNNs to demonstrate that the two models trained with different ways are sensitive 
to different kinds of features.
Besides this qualitative comparison, we also test the two kinds of CNNs on different transformed datasets to distinguish the difference of their preferred features. 

\begin{figure*}[ht]
\centering
\begin{tabular}{cccccc}
\includegraphics[width=2cm]{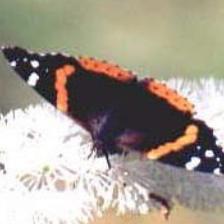} &
\includegraphics[width=2cm]{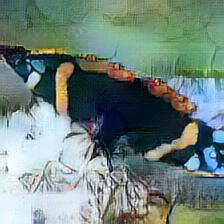} &
\includegraphics[width=2cm]{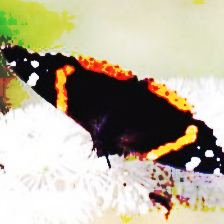}&
 \includegraphics[width=2cm]{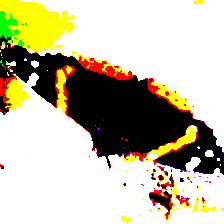}&
 \includegraphics[width=2cm]{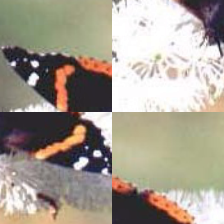} &
 \includegraphics[width=2cm]{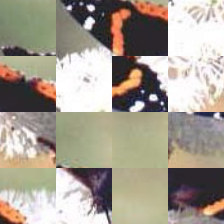}
\\
\includegraphics[width=2cm]{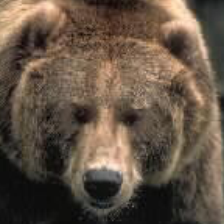} &
\includegraphics[width=2cm]{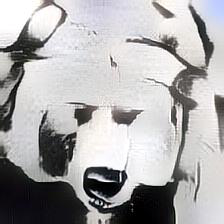} &
\includegraphics[width=2cm]{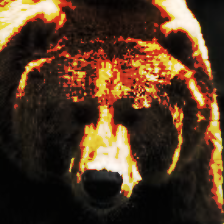}&
 \includegraphics[width=2cm]{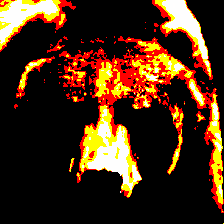}&
 \includegraphics[width=2cm]{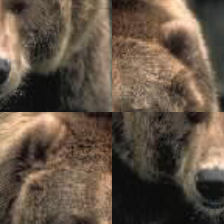} &
 \includegraphics[width=2cm]{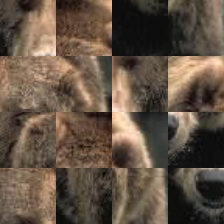} \\
(a) Original & (b) Stylized  & (c) Saturated 8  & (d) Saturated 1024  & (e) patch-shuffle 2  & (f) patch-shuffle 4
\end{tabular}
\caption{Visualization of three transformations. Original images are from Caltech-256. From left to right, original, stylized, saturation level as 8, 1024,  $2 \times 2$ patch-shuffling, $4 \times 4$ patch-shuffling.}
\label{fig:samples}
\end{figure*}

\subsection{Visualizing the salience maps}
A straightforward way of investigating the difference between AT-CNNs and CNNs is to visualize which group of pixels the network outputs are most sensitive to. Salience maps generated by Grad and its smoothed variant SmoothGrad are good candidates to show what features a model is sensitive to. 
We compare the salience maps between AT-CNNs and CNNs on clean images, and images under texture preserving and shape preserving distortions.  Extensive results can been seen in Section~\ref{sec:maps}.

As pointed by~\citet{smilkov2017smoothgrad}, sensitivity maps based on Grad method are often visually
noisy, highlighting that some pixels, to a human eye, seem randomly selected. SmoothGrad in Eq.~(\ref{eq:smoothgrad}), on the other hand, could reduce visual noise by averaging the gradient over the Gaussian perturbed images. Thus, we mainly report the salience maps produced by SmoothGrad, and the Grad visualization results are provided in the appendix. Note that the two visualization methods could help us draw a consistent conclusion on the difference between the two trained CNNs.

\subsection{Generalization on shape/texture  preserving distortions}
Besides visual inspection of sensitivity maps, we propose to measure the sensitivity of AT-CNNs and CNNs to different features by evaluating the performance degradation under several distortions that either preserves shapes or textures.
Intuitively, if one model relies on textures a lot, the performance would degrade severely if we destroy most of the textures while preserving other information, such as the shapes and other features. However, a perfect disentanglement of texture, shape and other feature information is impossible~\cite{gatys2015neural}. In this work, we mainly construct three kinds of image translations to achieve the shape or texture distortion, style-transfer, saturating and patch-shuffling operation. Some of the image samples are shown in Figure~\ref{fig:samples}. We also added three Fourier-filtered test set in the appendix. We now describe each of these transformations and their properties. 

Note that we conduct normal training or adversarial training on the original training sets, and then evaluate their generalizability over the transformed data. During the training, we never use the transformed datasets.  

\textbf{Stylizing}. \citet{geirhos2018imagenet} utilized style transfer~\cite{huang2017arbitrary} to generate images with conflicting shape and texture information to demonstrate the texture bias of ImageNet-trained standard CNNs. Following the same rationale, we utilize style transfer to destroy most of the textures while preserving the global shape structures in images, and build a stylized test dataset. Therefore, with similar generalization error, models capturing shapes better should also perform better on stylized test images than those biased towards textures. The style-transferred image samples are shown in  Figure~\ref{fig:samples}(b).

\textbf{Saturation}. Similar to~\cite{ding2018sensitivity}, we denote the saturation of the image $x$ by $x^p$, where $p$ indicates the saturation level ranging from $0$ to $\infty $. When $p = 2$, the saturation operation does not change the image. When $p \geq 2 $, increasing the saturation level will push the pixel values towards binarized ones, and $p = \infty$ leads to the pure binarization. Specifically, for each pixel of image $x$ with value $v\in [0,1]$, its corresponding saturated pixel of $x^p$ is defined as
$sign(2v -1) |2v -1|^{\frac{2}{p}}/2 + 1/2.$ 
One can observe that, from Figure~\ref{fig:samples}(c) and (d), increasing saturation level can gradually destroy some texture information while preserving most parts of the contour structures.

\begin{table*}[ht!]
\caption{Accuracy and robustness of all the trained models. Robustness is measured against the PGD attack with bounded  $l_{\infty}$ norm. Details are listed in the appendix. Note that underfitting CNNs have similar generalization performance with some of the AT-CNNs on clean images. }
\label{tab:Model}
\begin{center}
\begin{tabular}{|c|c|c|c|c|c|c|}
\hline
 & \multicolumn{2}{l|}{CIFAR10} & \multicolumn{2}{l|}{TinyImageNet} & \multicolumn{2}{l|}{Caltech 256} \\ \cline{2-7} 
                  & Accuracy        &  Robustness     & Accuracy         &      Robustness  &     Accuracy     &   Robustness        \\ \hline
  PGD-inf: 8               &  86.27       &  \textbf{44.81 }         &      54.42     &   14.25        &   66.41        &    31.16       \\ \hline
  PGD-inf: 4               &  89.17       &  30.85          &       61.85    &  6.87         &        72.22   &     20.10      \\ \hline
  PGD-inf: 2               &   91.4        &  39.11         &       67.06    &  1.66         &         76.51  &     7.51      \\ \hline
  PGD-inf: 1               &  93.40         &  7.53        &       69.42    &      0.18     &       79.11    &      1.70     \\ \hline
  PGD-L2: 12               &     85.79      & 34.61          &        53.44   & \textbf{ 14.80}         &       65.54    &        \textbf{31.36}   \\ \hline
  PGD-L2: 8               &   88.01       &   26.88        &         58.21 &    10.03       &        69.75   &        26.19    \\ \hline
  PGD-L2: 4               &  90.77         &    13.19       &         64.24  &       3.61   &        74.12   &      14.33     \\ \hline
  FGSM: 8              &    84.90       &     34.25      &           66.21   &   0.01       &       70.88    &       20.02    \\ \hline
  FGSM: 4              &   88.13         &    25.08       &           63.43 &     0.13      &       73.91    &         15.16  \\ \hline
  Normal              &      \textbf{94.52}     &      0     &         \textbf{72.02}  &  0.01         &        \textbf{83.32}   &    0       \\ \hline
  Underfit              &     86.79     &   0        &          60.05 &     0.01      &      69.04     &        0   \\ \hline
\end{tabular}
\end{center}
\end{table*}

\textbf{Patch-Shuffling}. To destroy long-range shape information, we split images into $k \times k$ small patches and randomly rearranging the order of these patches, with $k \in \{2,4,8 \}$. Favorably, this operation preserves most of the texture information and destroys most of the shape information. The patch-shuffled image samples are showed in Figure~\ref{fig:samples}(e), (f).  Note that as $k$ increasing, more information of the original image is lost, especially for images with low resolution.

\section{Experiments and analysis}
\label{sec:exp}

\paragraph{Experiments setup}
We describe the experiment setup to evaluate the performance of AT-CNNs and standard CNNs in data distributions manipulated by above-mentioned operations. 
We conduct experiments on three datasets. CIFAR-10, Tiny ImageNet and Caltech-256~\cite{griffin2007caltech}. Note that we do not create the style-transferred and patch-shuffled test set for CIFAR-10 due to its limited resolution. 

When training on CIFAR-10, we use the ResNet-18 model~\cite{he2016deep, he2016identity}; for data augmentation, we perform zero paddings with width as 4, horizontal flip and random crop. 

Tiny ImageNet has 200 classes of objects. Each class has 500 training images, 50 validation images, and 50 test images. 
All images from Tiny ImageNet are of size $64 \times 64$. We re-scale them to $224 \times 224$ and perform random horizontal flip and per-image standardization as data augmentation.

\begin{figure*}[ht!]
\centering
\begin{tabular}{cc}
\includegraphics[width=8cm]{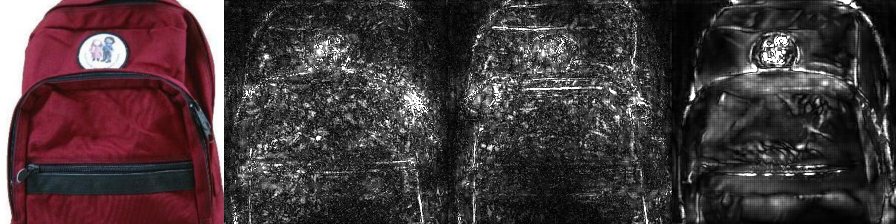} & \includegraphics[width=8cm]{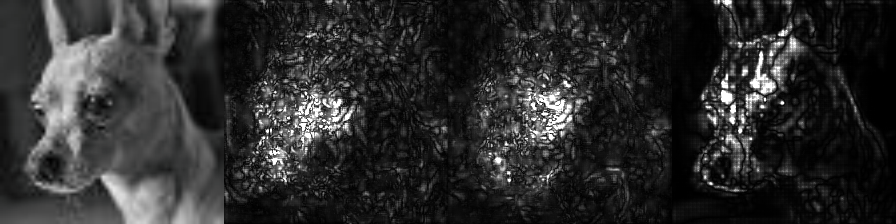}  \\
\includegraphics[width=8cm]{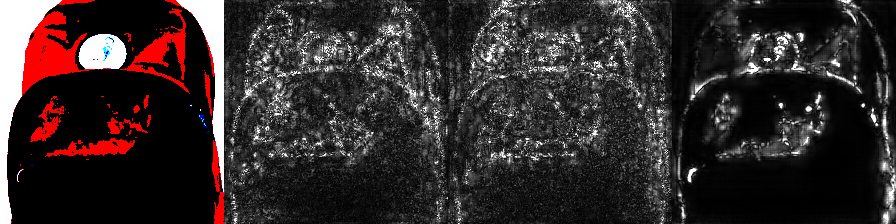} & 
\includegraphics[width=8cm]{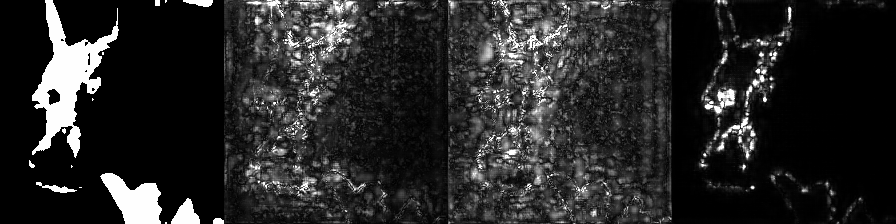}  \\
\includegraphics[width=8cm]{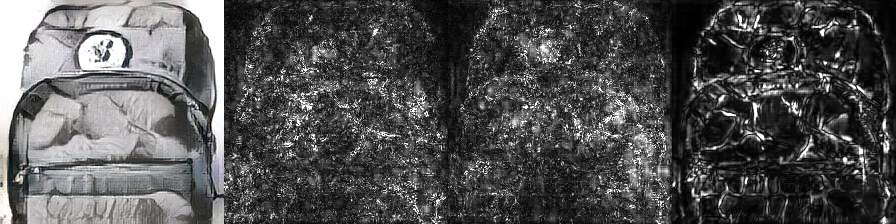} & 
\includegraphics[width=8cm]{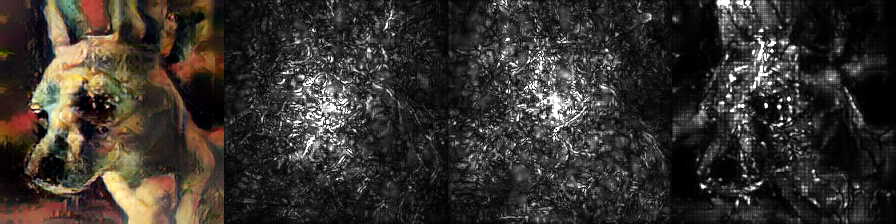}  \\
(a) Images from Caltech-256 & (b) Images from Tiny ImageNet \\
\end{tabular}
\caption{Sensitivity maps based on SmoothGrad~\cite{smilkov2017smoothgrad} of three models on images under saturation, and stylizing. From top to bottom, Original, Saturation 1024 and Stylizing. For each group of images, from left to right, original image, sensitivity maps of standard CNN, underfitting CNN and PGD-$l_{\infty}$ AT-CNN. }
\label{fig:salience}
\end{figure*}

Caltech-256~\cite{griffin2007caltech} consists of 257 object categories containing a total of 30607 images. Resolution of images from Caltech is much higher compared with the above two datasets. We manually split $20\%$ of images as the test set.  We perform re-scaling and random cropping following~\cite{he2016deep}. For both Tiny ImageNet and Caltech-256, we use ResNet-18 model as the network architecture.

\paragraph{Compared models, their generalization and robustness.} For all above three datasets, we train three types of AT-CNNs, they mainly differ in the way of generating adversarial examples: FGSM, PGD with bounded $l_{\infty}$ norm and PGD with bounded $l_2$ norm, and for each attack method we train several models under different attack strengths. Details are listed in the appendix.  To understand whether the difference of performance degradation  for AT-CNNs and standard CNNs is due to the poor generalization~\cite{schmidt2018adversarially, tsipras2018robustness} of adversarial training, we also compare the AT-CNNs with an underfitting CNN (trained over clean data) with similar generalization performance as AT-CNNs. We train 11 models on each dataset. Their generalization performance on clean data, and robustness measured by PGD attack are shown in Table ~\ref{tab:Model}.

\subsection{Visualization results}
\label{sec:maps}
To investigate what features of an input image AT-CNNs and normal CNNs are most sensitive to, we generate sensitivity maps using SmoothGrad~\cite{smilkov2017smoothgrad} on clean images, saturated images, and stylized images.  The visualization results are presented in Figure~\ref{fig:salience}. 

We can easily observe that the salience maps of AT-CNNs are much more sparse and mainly focus on contours of each object on all kinds of images, including the clean, saturated and stylized ones.  Differently, sensitivity maps of standard CNNs are more noisy, and less biased towards the shapes of objects. This is consistent with the findings in~\cite{geirhos2018imagenet}. 

Particularly, in the second row of Figure~\ref{fig:salience}, sensitivity maps of normal CNNs of the ``dog'' class are still noisy even when the input saturated image are nearly binarized. On the other hand, after adversarial training, the models successfully capture the shape information of the object, providing a more interpretable prediction. 

For stylized images shown in the third row of Figure~\ref{fig:salience}, even with dramatically  changed textures after style transfer, AT-CNNs can still be able to focus the shapes of original object, while standard CNNs totally fail. 

Due to the limited space, we provide more visualization results (including the sensitivity maps generated by Grad method) in appendix. 

\subsection{Generalization performance on transformed data}
\label{sec:acc}
In this part, we mainly show generalization performance of AT-CCNs and normal CNNs on either shape or texture preserving distorted image datasets. This could help us to understand how different  that the two types of models are biased in a quantitative way. 

For all experimental results below, besides the top-1 accuracy, we also report an ``\emph{accuracy on correctly classified images}''.  This accuracy is measured by first selecting the images from the clean test set that is being correctly classified, then measuring the accuracy of transformed images from these correctly classified ones.

\begin{figure*}[h!]
\label{fig:style-samples}
\centering
\begin{tabular}{cccc}
\includegraphics[width=3.8cm]{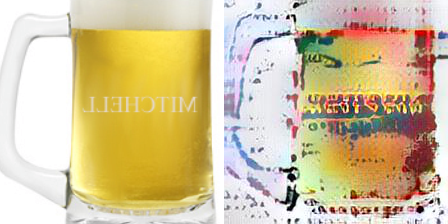} & \includegraphics[width=3.8cm]{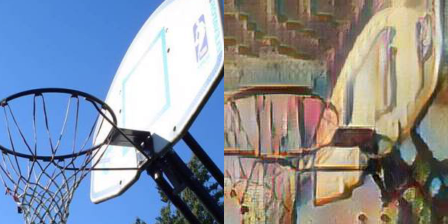}  & \includegraphics[width=3.8cm]{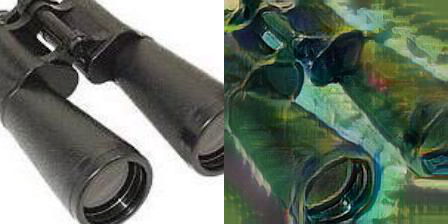} &  \includegraphics[width=3.8cm]{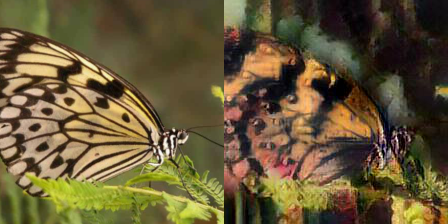} \\
\includegraphics[width=3.8cm]{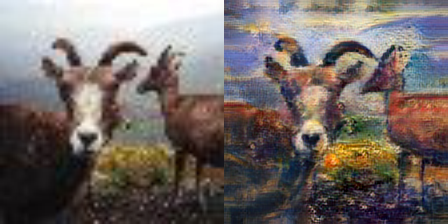} & \includegraphics[width=3.8cm]{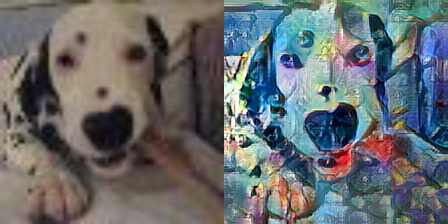} &
\includegraphics[width=3.8cm]{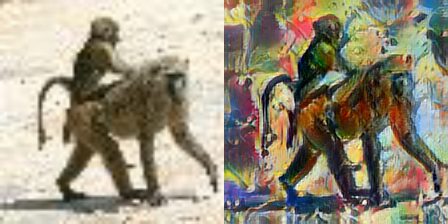} &  \includegraphics[width=3.8cm]{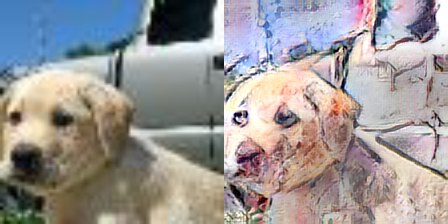}
\end{tabular}
\vspace{-0.1cm}
\caption{Visualization of images from style-transferred test set. Applying AdaIn~\cite{huang2017arbitrary} style transfer distorts local textures of original images, while the global shape structure is retained. The first row are images from Caltech-256, and the second row are images from Tiny ImageNet. }
\vskip -0.2in
\end{figure*}

\begin{table*}[h!]
\caption{``Accuracy on correctly classified images''  for different models on stylized test set. The columns named ``Caltech-256'' and ``TinyImageNet'' show the generalization of different models on the clean test set.}
\label{tab:style-acc}
\vskip -0.15in
\begin{center}
\begin{small}
\begin{sc}
\begin{tabular}{lcccr}
\toprule
dataset &Caltech-256 & Stylized Caltech-256 & TinyImageNet & Stylized TinyImageNet \\
\midrule
Standard    &\textbf{ 83.32}  &16.83 & \textbf{72.02} &  7.25 \\
Underfit & 69.04   &9.75                  & 60.35 & 7.16   \\
PGD-$l_{\infty}$: 8 & 66.41    & 19.75      & 54.42 & 18.81   \\
PGD-$l_{\infty}$: 4 & 72.22    & 21.10      & 61.85 & 20.51  \\
PGD-$l_{\infty}$: 2 & 76.51    & 21.89      & 67.06 & 19.25  \\
PGD-$l_{\infty}$: 1 & 79.11    & 22.07      & 69.42 & 18.31  \\
PGD-$l_{2}$: 12     & 65.24    & 20.14      & 53.44 & 19.33  \\
PGD-$l_{2}$: 8      & 69.75    & 21.62      & 58.21 & 20.42  \\
PGD-$l_{2}$: 4      & 74.12    & \textbf{22.53}  & 64.24 & \textbf{21.05}  \\
FGSM: 8             & 70.88    & 21.23      & 66.21 & 15.07 \\
FGSM: 4             & 73.91    & 21.99      &  63.43 & 20.22 \\
\bottomrule
\end{tabular}
\end{sc}
\end{small}
\end{center}
\vskip -0.1in
\end{table*}

\begin{figure*}[ht!]
\centering
\begin{tabular}{c}
\includegraphics[width=16cm]{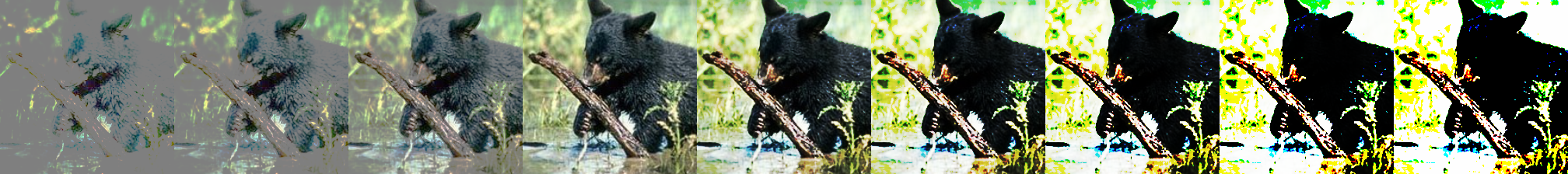} 
\end{tabular}

\caption{Illustration of how varying saturation changes the appearance of the image. From left to right, saturation level 0.25, 0.5, 1, 2 (original image), 4, 8, 16, 64, 1024. Increasing saturation level pushes pixels towards 0 or 1, which preserves most of the shape while wiping most of the textures. Decreasing saturation level pushes all pixels to ${1}/{2}$.}
\vskip -0.1in
\label{fig:sat-increase}
\end{figure*}

\subsubsection{Stylizing}
Following \citet{geirhos2018imagenet}, we generate stylized version of test set for Caltech-256 and Tiny ImageNet. 

We report the ``accuracy on correctly classified images'' of all the trained models on stylized test set in Table~\ref{tab:style-acc}. Compared with standard CNNs, though with a lower accuracy on original test images, AT-CNNs achieve higher accuracy on stylized ones with textures being dramatically changed. The comparison quantitatively shows that AT-CNNs tend to be more invariant with respect to local textures.

\begin{figure*}[ht!]
\centering
\begin{tabular}{cc}
\includegraphics[width=0.8\columnwidth]{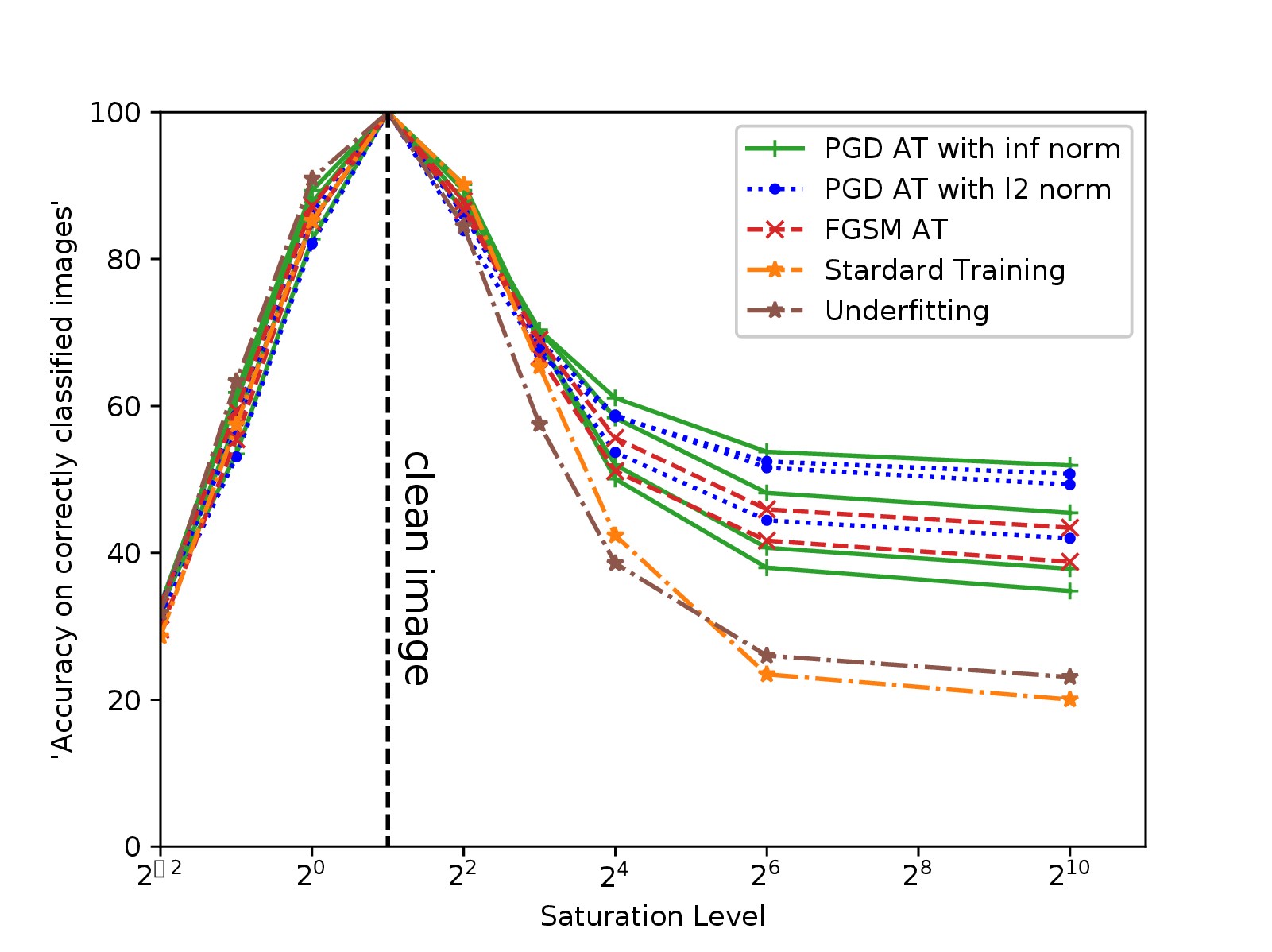} &
\includegraphics[width=0.8\columnwidth]{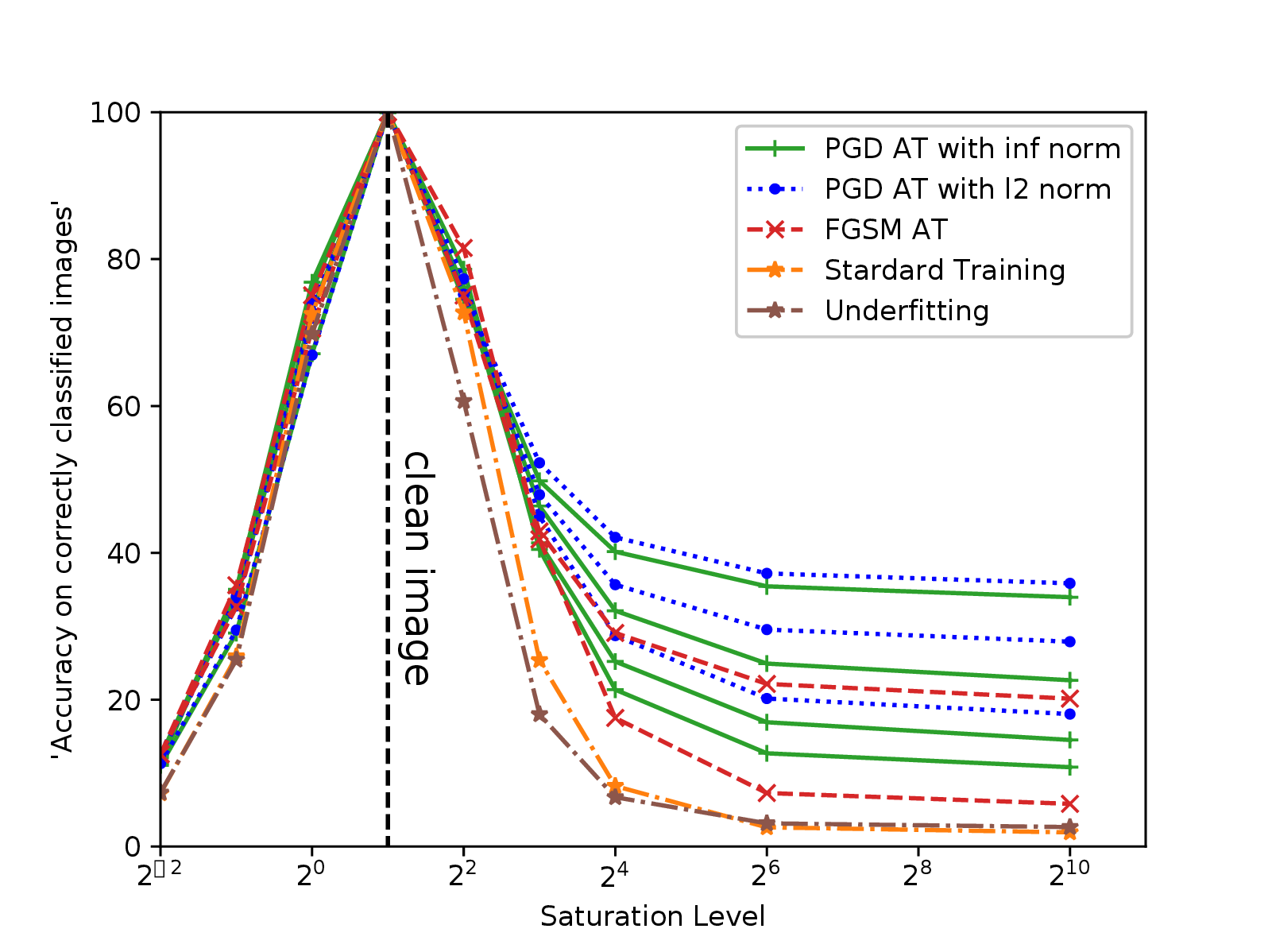} \\
(a) Caltech-256 & (b)  Tiny ImageNet
\end{tabular}
\vskip -0.1in
\caption{``Accuracy on correctly classified images'' for different models on saturated Caltech-256 and Tiny ImageNet with respect to different saturation levels. Note that in the plot, there are several curves with same color and line type shown for each adversarial training method, PGD and FGSM-based, those of which with larger perturbation achieves better robustness for most of the cases. Detailed results are list in the appendix.}
\label{fig:sat-acc}
\end{figure*}

\begin{figure*}[ht!]
\centering
\begin{tabular}{cccc}
\includegraphics[width=3cm]{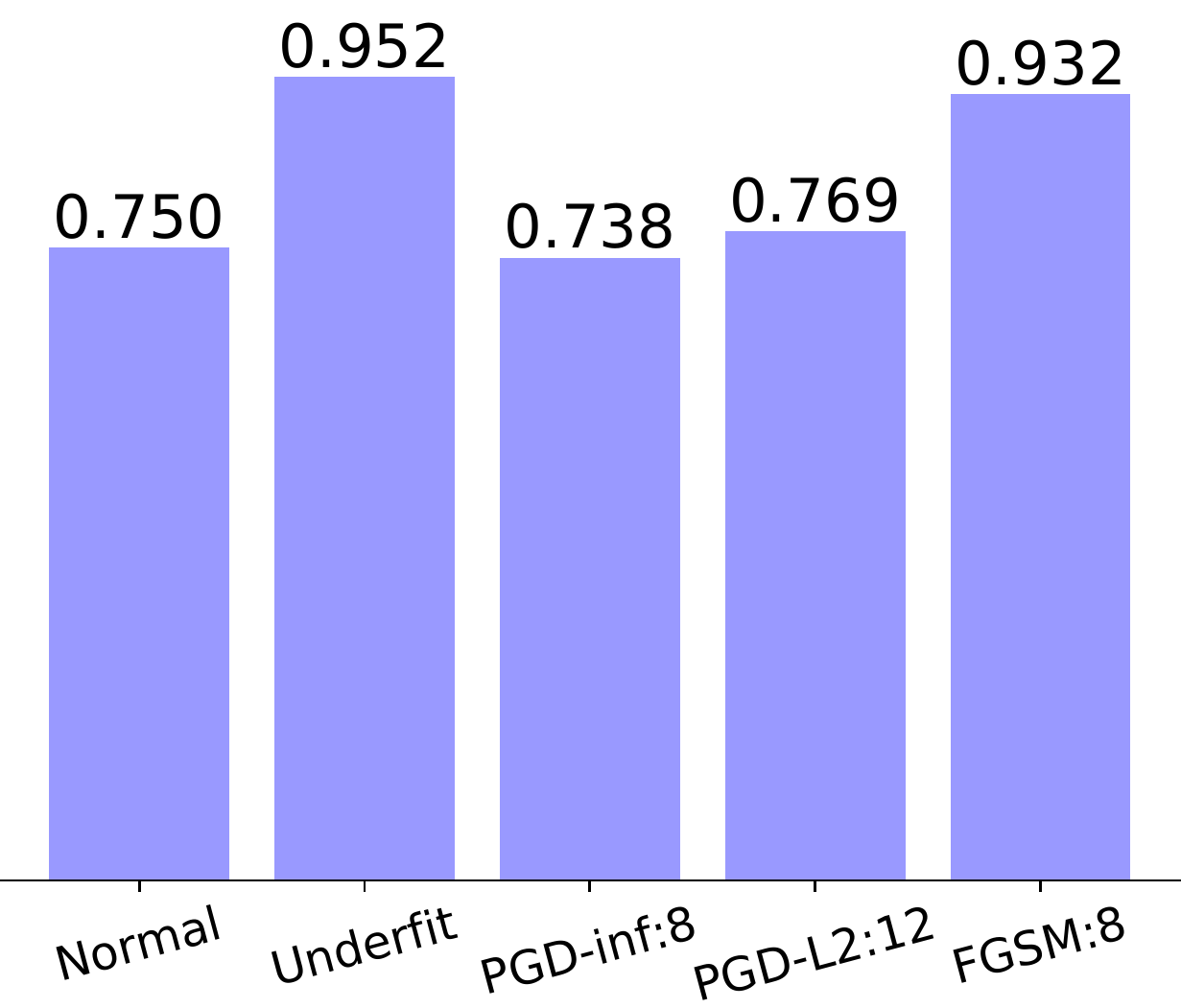} &
\includegraphics[width=3cm]{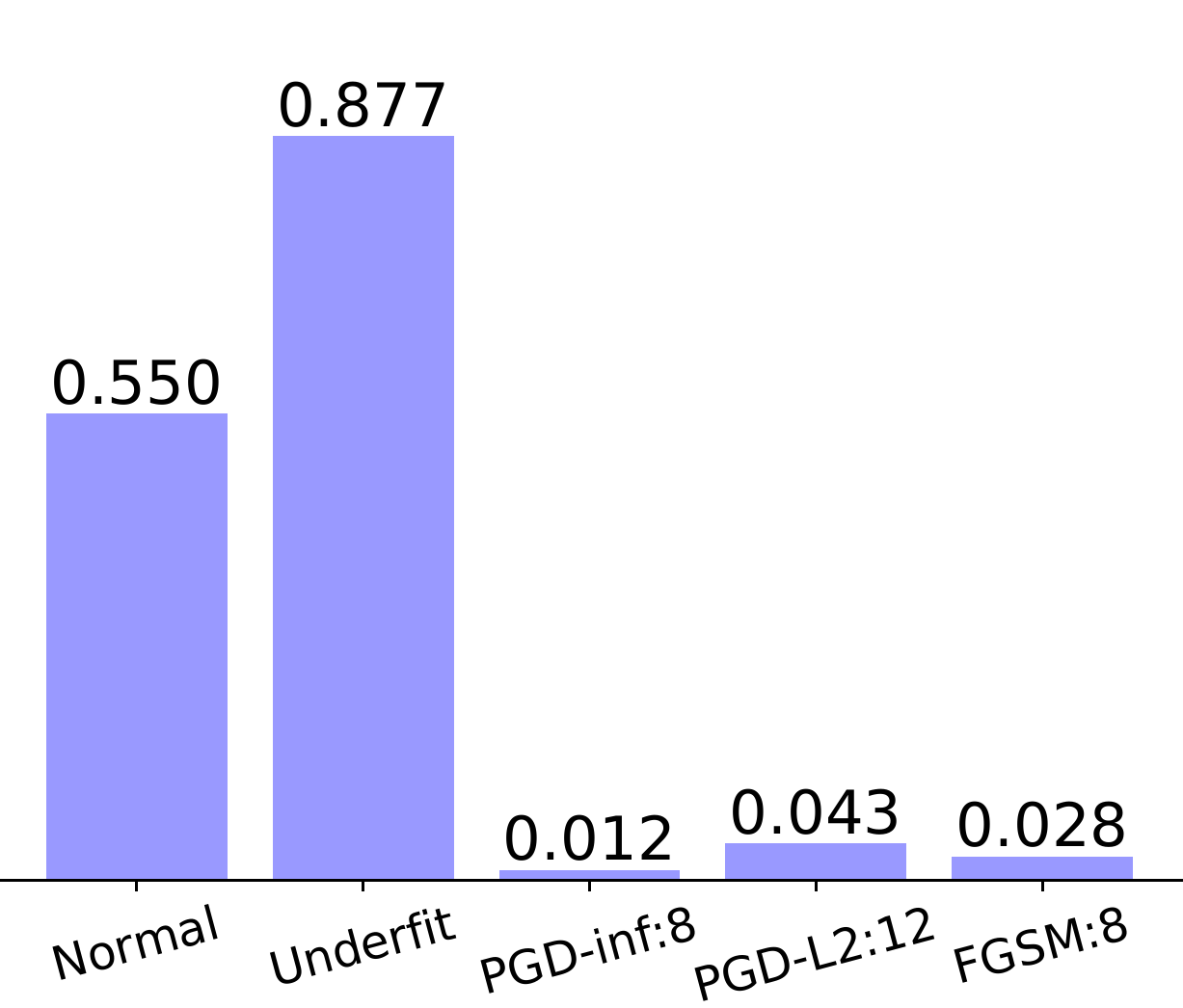} &  
\includegraphics[width=3cm]{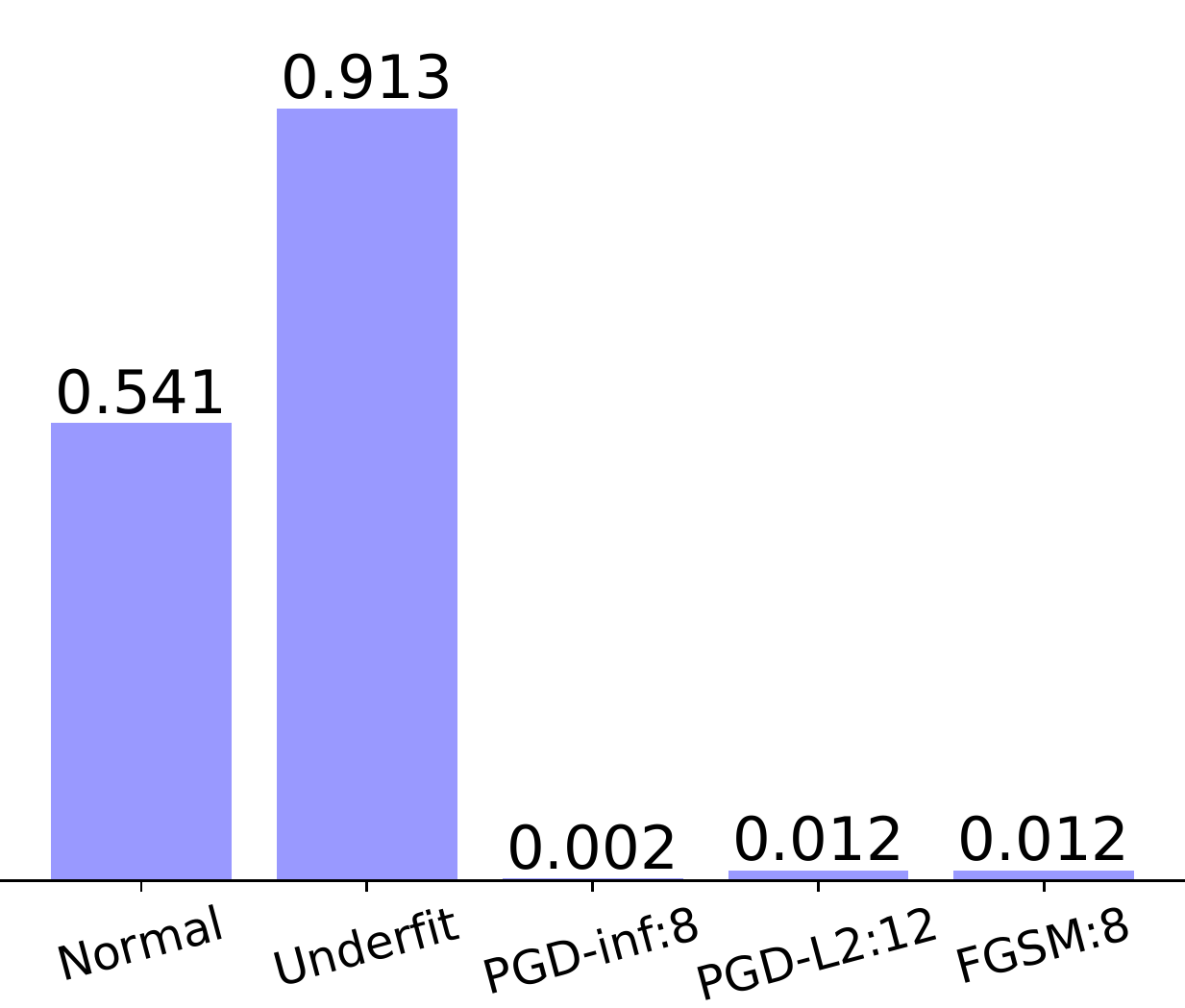} &
\includegraphics[width=3cm]{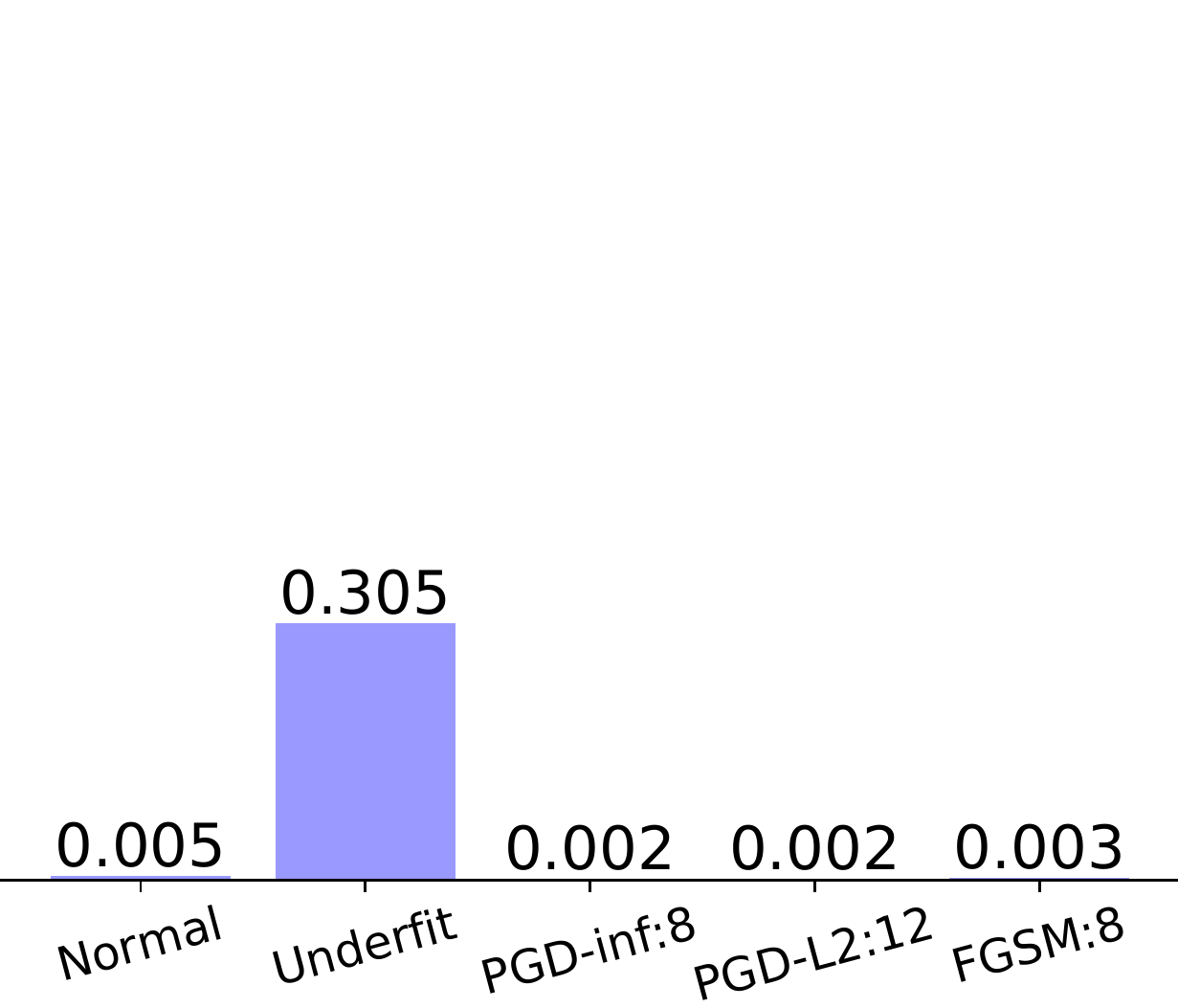} 
\\
\includegraphics[width=3cm]{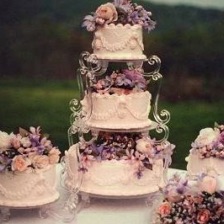} &
\includegraphics[width=3cm]{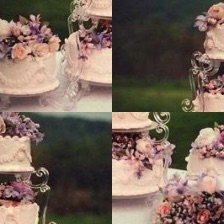} &  
\includegraphics[width=3cm]{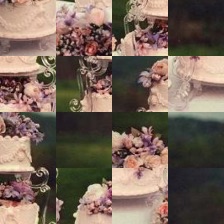} &
\includegraphics[width=3cm]{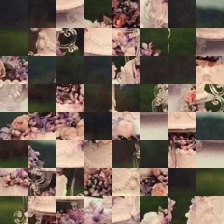} 
\\
(a) Original Image & (b) Patch-Shuffle 2  & (c) Patch-Shuffle 4  & (d) Patch-Shuffle 8
\end{tabular}
\vskip -0.1in
\caption{Visualization of patch-shuffling transformation. The first row shows  probability of ``cake'' assigned by different models.}
\label{fig:patch-shuffle}
\end{figure*}

\begin{figure*}[ht!]
\centering
\begin{tabular}{cc}
\includegraphics[width=0.8\columnwidth]{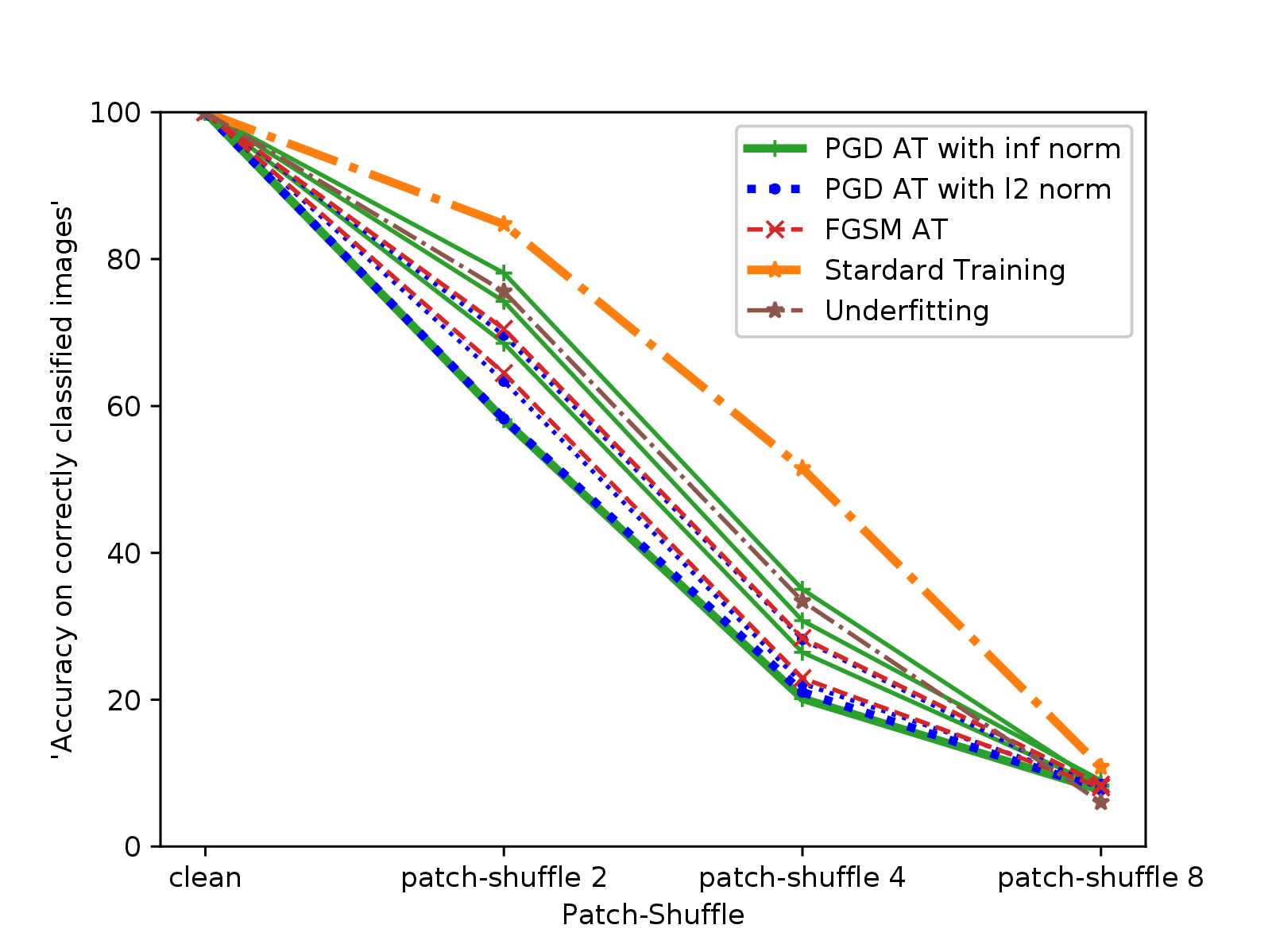} &
\includegraphics[width=0.8\columnwidth]{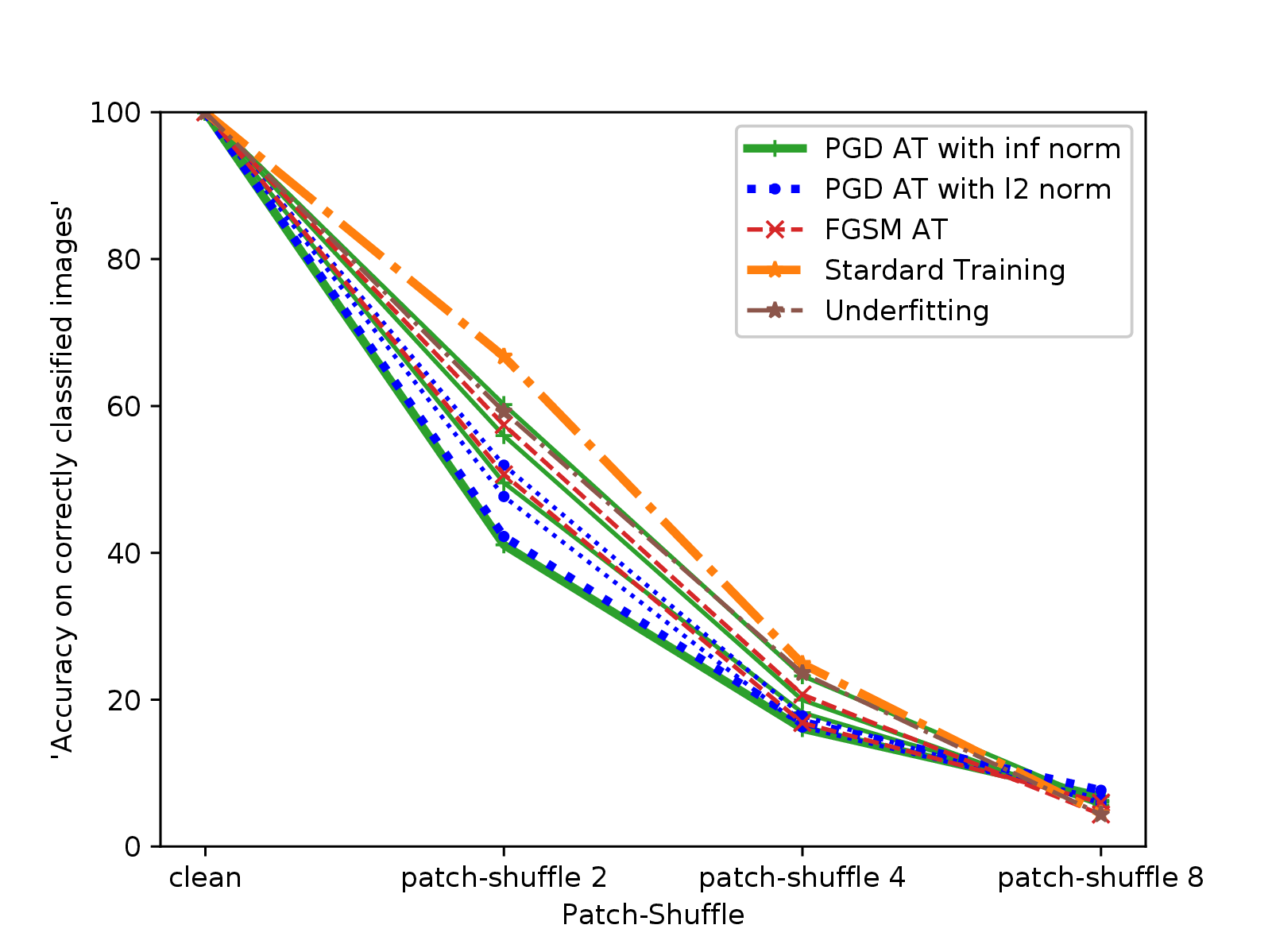} \\
(a) Caltech-256 & (b) Tiny ImageNet
\end{tabular}
\vskip -0.1in
\caption{``Accuracy on correctly classified images'' for different models on patch-shuffled Tiny ImageNet and Caltech-256 with different splitting numbers. Detailed results are listed in the appendix.}
\label{fig:patch-acc}
\end{figure*}

\subsubsection{Saturation}
We use the saturation operation to manipulate the images, and show the how increasing saturation levels affects the accuracy of models trained in different ways.

In Figure~\ref{fig:sat-increase}, we visualize images with varying saturation levels. It can be easily observed that increasing saturation levels pushes images more ``binnarized'', where some textures are wiped out, but produces sharper edges and preserving shape information. When saturation level is smaller than $2$, i.e. clean image, it pushes all the pixels towards $1/2$ and nearly all the  information is lost, and $p = 0$ leads to a totally gray image with constant pixel value.

We measure the ``accuracy on correctly classified images'' for all the trained models, and show them in Figure~\ref{fig:sat-acc}.  We can observe that with the increasing level of saturation, more texture information is lost. Favorably, adversarially trained models exhibit a much less sensitivity to this texture loss, still obtaining a high classification accuracy.  The results indicate that AT-CNNs are more robust to ``saturation'' or ``binarizing'' operations, which may demonstrate that the prediction capability of AT-CNNs relies less on texture and more on shapes. Results on CIFAR-10 tells the same story, as presented in appendix due to the limited space.

Additionally, in our experiments, for each adversarial training approach, either PGD or FGSM based, AT-CNNs with higher robustness towards PGD adversary are more invariant to the increasing of the saturation level and texture loss. On the other hand, adversarial training with higher robustness typically ruin the generalization over the clean dataset. Our finding also supports the claim ``robustness maybe at odds with accuracy''~\cite{tsipras2018robustness}.     

When decreasing the saturation level, all models have similar degree of performance degradation, indicating that AT-CNNs are not robust to all kinds of image distortions. They tend to be more robust for fixed types of distortions. We leave the further investigation regarding this issue as future work.

\subsubsection{Patch-Shuffling}
Stylizing and saturation operation aim at changing or removing the texture information of original images, while preserving the features of shapes and edges. In order to test the different bias of AT-CNN and standard CNN in the other way around, we shatter the shape and edge information by splitting the images into $k \times k $ patches and then randomly shuffling them. This operation could still maintains the local textures if $k$ is not too large.

Figure~\ref{fig:patch-shuffle} shows one example of patch-shuffled images under different numbers of splitting. The first row shows the probabilities assigned by different models to the ground truth class of the original image.  Obviously, after random shuffling, the shapes and edge features are destroyed dramatically, the prediction probability of the adverarially trained CNNs drops significantly, while the normal CNNs still maintains a high confidence over the ground truth class. This reveals AT-CNNs are more baised towards shapes and edges than normally trained ones.
 
 Moreover, Figure \ref{fig:patch-acc} depicts the ``
 accuracy of correctly classified images'' for all the models measured on ``Patch-shuffled'' test set with increasing number of splitting pieces. AT-CNNs,  especially trained against with a stronger attack are more sensitive to ``Patch-shuffling'' operations in most of our experiments.
 
 Note that under ``Patch-shuffle 8'' operation, all models have similar ``
 accuracy of correctly classified images'', which is largely due to the severe information loss.  
 Also note that this accuracy of all models on Tiny ImageNet shown in \ref{fig:patch-acc}(a) is mush lower than that on Caltech-256 in \ref{fig:patch-acc}(b). That is, under ``Patch-shuffle 1'', normally trained CNN has an accuracy of $84.76 \%$ on Caltech-256, while only $66.73 \%$ on Tiny ImageNet. This mainly origins from the limited resolution of Tiny ImageNet, since ``Patch-Shuffle'' operation on  low-resolution images  destroys more useful features than those with higher resolution.

\section{Related work and discussion}
\label{sec:discussion}
\textbf{Interpreting AT-CNNs.} Recently there are some relevant findings indicating that AT-CNNs learn fundamentally different feature representations than standard classifiers. \citet{tsipras2018robustness} showed that sensitivity maps of AT-CNNs in the input space align well with human perception. Additionally, 
by visualizing large-$\varepsilon$ adversarial examples against AT-CNNs,  it can be observed that the adversarial examples could capture salient data characteristics of a different class, which appear semantically similar to the images of the different class.  \citet{dong2017towards} leveraged adversarial training to produce a more interpretable representation by visualizing active neurons. 
Compared with \citet{tsipras2018robustness} and \citet{dong2017towards}, we have conducted a more systematical investigation for interpreting AT-CNNs. We construct three types of image transformation that can largely change the textures while preserving shape information (i.e. stylizing and saturation),  or shatter the shape/edge features while keeping the local textures (i.e. patch-shuffling). Evaluating the generalization of AT-CNNs over these designed datasets provides a \emph{quantitative} way to verify and interpret their strong shape-bias compared with normal CNNs.



\textbf{Insights for defensing adversarial examples.} Based on our investigation over the AT-CNNs, we find that the robustness towards adversarial examples is correlated with the capability of capturing long-range features like shapes or contours. This naturally raises the question: 
\emph{whether any other models that can capture more global features or with more texture invariance could lead to  more robustness to adversarial examples, even without adversarial training?} This might provide us some insights on designing new network architecture or new strategies for enhancing the bias towards long-range features. Some recent works turn out partially answering this question.  \citep{xie2018feature} enhanced standard CNNs with non-local blocks inspired from \cite{wang2018non, vaswani2017attention} which capture long-range dependencies in a data-dependent manner, and when combined with adversarial training, their networks achieved state-of-the-art adversarial robustness on ImageNet. \cite{luo2018random} destroyed some of the local connection of standard CNNs by randomly select a set of neurons and remove them from the network before training, and thus forcing the CNNs to less focus on local texture features. With this design, they achieved improved black-box robustness.

\begin{figure}[ht!]
\centering
\includegraphics[width=8cm]{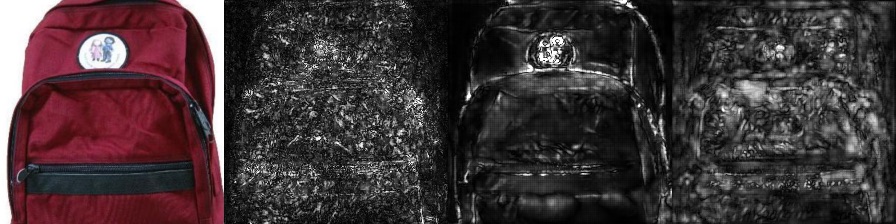}

\caption{Sensitivity maps based on SmoothGrad~\cite{smilkov2017smoothgrad} of three models. From left to right, original image, sensitivity maps of standard CNN, PGD-$l_{\infty}$ AT-CNN and ST-AT-CNN. }
\vskip -0.1in
\label{fig:stadv}
\end{figure}

\textbf{Adversarial training with other types of attacks.} In this work, we mainly interpret the AT-CNNs based on norm-constrained perturbation over the original images. It is worthy of noting that the difference between normally trained and adversarially trained CNNs may highly depends on the type of adversaries.   Models trained against spatially-transformed adversary~\cite{xiao2018spatially}, denoted as ST-ST-CNNs,  have similar robustness towards PGD attack with standard models, and their salience maps are still quite different as shown in Figure~\ref{fig:stadv}. Also the average distance between salience maps is close to that of standard CNN, which is much higher than that of PGD-AT-CNN. 
There exists a variety of generalized types of attacks, $x_{\textrm{adv}} = G(x; w)$ parameterized by $w$, such as spatially transformed~\cite{xiao2018spatially} and GAN-based adversarial examples~\cite{song2018constructing}. We leave interpreting the AT-CNNs based on these generalized types of attacks  as future work.  

\section{Conclusion}
\label{sec:con}
From both qualitative and quantitative perspectives, we have implemented a systematic study on interpreting the adversarially trained convolutional neural networks. Through constructing distorted test sets either preserving shapes or local textures, we compare the sensitivity maps of AT-CNNs and normal CNNs on the clean, stylized and saturated images, which visually demonstrates that AT-CNNs are more biased towards global structures, such as shapes and edges. More importantly, we evaluate the generalization performance of the two models on the three constructed datasets, stylized, saturated and patch-shuffled ones. The results clearly indicate that AT-CNNs are less sensitive to the texture distortion and focus more on shape information, while the normally trained CNNs the other way around. 

Understanding what a model has learned is an essential topic in both machine learning and computer vision. The strategies we propose can also be extended to interpret other neural networks, such as models for object detection and semantic segmentation.


\section*{Acknowledgement}
This  work  is  supported  by  National  Natural  Science  Foundation  of  China  (No.61806009),  Beijing Natural  Science  Foundation  (No.4184090), Beijing Academy of Artificial Intelligence (BAAI) and Intelligent  Manufacturing  Action  Plan  of  Industrial  Solid Foundation  Program  (No.JCKY2018204C004). We also appreciate insightful discussions with Dinghuai Zhang and Dr. Lei Wu.

\bibliography{atcnn}
\bibliographystyle{icml2019}

\newpage

\appendix

\section{Experiment Setup}

\subsection{Models}
\begin{itemize}
    \item \textbf{CIFAR-10}. We train a standard ResNet-18 \cite{he2016deep} architecture, it has 4 groups of residual layers with filter sizes (64, 128, 256, 512) and 2 residual units.
    
    \item \textbf{Caltech-256 \& Tiny ImageNet}. We use a ResNet-18 architecture using the code from pytorch\cite{paszke2017automatic}. Note that for models on Caltech-256 \& Tiny ImageNet, we initialize them using ImageNet\cite{deng2009imagenet} pre-trained weighs provided by pytorch.
\end{itemize}
We evaluate the robustness of all our models using a $l_{\infty}$ projected gradient descent adversary  with $\epsilon = 8/255$, step size = 2 and number of iterations as 40.

\subsection{Adversarial Training}
We perform 9 types of adversarial training on each of the dataset. 7 of the 9 kinds of adversarial training are against a projected gradient descent (PGD) adversary\cite{madry2017towards}, the other 2 are against FGSM adversary\cite{goodfellow2014explaining}.

\subsubsection{Train against a projected gradient descent (PGD) adversary}
We list value of $\epsilon$ for adversarial training of each dataset and $l_p$-norm. In all settings, PGD runs 20 iterations.
\begin{itemize}
    \item \textbf{$l_{\infty}$-norm bounded adversary}. For all of the three data set, pixel vaules range from 0~1, we train 4 adversarially trained CNNs with $\epsilon \in \{1/255, 2/255, 4/255, 8/255 \}$,  these four models are denoted as PGD-inf:1, 2, 4, 8 respectively, and steps size as 1/255, 1/255, 2/255, 4/255. 
    \item \textbf{$l_{2}$-norm bounded adversary}. For Caltech-256 \& Tiny ImageNet, the input size for our model is $224 \times 224$, we train three adversarially trained CNNs with $\epsilon \in \{4, 8, 12\}$, and these four models are denoted as PGD-l2: 4, 8, 12 respectively. Step sizes for these three models are 2/255, 4/255, 6/255. For CIFAR-10, where images are of size $32 \times 32$, the three adversarially trained CNNs have $\epsilon \in \{ 4/10, 8/10, 12/10 \}$, but they are denoted in the same way and have the same step size as that in Caltech-256 \& Tiny ImageNet.

\subsubsection{Train against a FGSM adversary}
$\epsilon$ for these two adversarially trained CNNs are $\epsilon \in \{4, 8 \}$, and they are denoted as FGSM 4, 8 respectively.
\end{itemize}

\section{Style-transferred test set}
Following \cite{geirhos2018imagenet} we construct stylized test set for Caltech-256 and Tiny ImageNet by applying the AdaIn style transfer\cite{huang2017arbitrary} with a stylization coefficient of $\alpha = 1.0$ to every test image with the style of a randomly selected painting from \footnote{\url{https://www.kaggle.com/c/painter-by-numbers/}}Kaggle's \textit{Painter by numbers} dataset. we used source code provided by\cite{geirhos2018imagenet}. 

\section{Experiments on Fourier-filtered datasets}

\cite{jo2017measuring} showed deep neural networks tend to learn surface
statistical regularities as opposed to high-level abstractions. Following them, we test the performance of different trained CNNs on the high-pass and low-pass filtered dataset to show their tendencies.

\subsection{Fourier filtering setup}
Following \cite{jo2017measuring} We construct three types of Fourier filtered version of test set.
\begin{itemize}
    \item \textbf{The low frequency filtered version}. We use a radial mask in the Fourier domain to set higher frequency modes to zero.(low-pass filtering)
    \item \textbf{The high frequency filtered version}. We use a radial mask in the Fourier domain to preserve only the higher frequency modes.(high-pass filtering)
    \item \textbf{The random filtered version}. We use a random mask in the Fourier domain to set each mode to 0 with probability $p$ uniformly. The random mask is generated on the fly during the test.
    
\end{itemize}
\subsection{Results}
We measure generalization performance (accuracy on correctly classified images) of each model on these three filtered datasets from Caltech-256, results are listed in Table~\ref{tab:four-acc}. AT-CNNs performs better on Low-pass filtered dataset and worse on High-pass filtered dataset. Results indicate that AT-CNNs make their predictions depend more on low-frequency information. This finding is consistent with our conclusions since local features such as textures are often considered as high-frequency information, and shapes and contours are more like low-frequency.

\begin{table*}[h!]
\caption{``Accuracy on correctly classified images''  for different models on three Fourier-filtered Caltech-256 test sets.}
\label{tab:four-acc}
\vskip -0.15in
\begin{center}
\begin{small}
\begin{sc}
\begin{tabular}{lccc}
\toprule
Data set & The low frequency filtered version & The high frequency filtered version & The random filtered version  \\
\midrule
Standard    &  15.8  & 16.5  & 73.5  \\
Underfit    & 14.5   & 17.6  & 62.2  \\
PGD-$l_{\infty}$: & 71.1 & 3.6  & 73.4  \\

\bottomrule
\end{tabular}
\end{sc}
\end{small}
\end{center}
\vskip -0.1in
\end{table*}

\section{Detailed results}

We  the detailed results for our quantitative experiments here. Table \ref{tab:tiny-sat-all}, \ref{tab:cal-sat-all}, \ref{tab:cifar-sat-all} show the results of each models on test set with different saturation levels.
Table \ref{tab:path-tiny}, \ref{tab:patch-cal} list all the results of each models on test set after different path-shuffling operations.

\begin{table*}[h!]
\caption{``Accuracy on correctly classified images''  for different models on saturated Caltech-256 test set. It is easily observed AT-CNNs are much more robust to increasing saturation levels on Caltech-256.}
\label{tab:cal-sat-all}
\vskip -0.15in
\begin{center}
\begin{small}
\begin{sc}
\begin{tabular}{lcccccccc}
\toprule
Saturaion level & 0.25 & 0.5 & 1 & 4 & 8 & 16 & 64 & 1024 \\
\midrule
Standard    & 28.62  & 57.45 & 85.20  & 90.13 &  65.37 &  42.37 & 23.45 &  20.03  \\
Underfit & 31.84   & \textbf{63.36}  & \textbf{90.96} &  84.51  & 57.51 & 38.58 & 26.00 & 23.08                  \\
PGD-$l_{\infty}$: 8 & 32.84    & 53.47  & 82.72 & 86.45  &  \textbf{70.33}  & \textbf{61.09}  & \textbf{53.76}  & \textbf{51.91}   \\
PGD-$l_{\infty}$: 4 & 31.99    & 57.74      & 85.18 & 87.95 & \textbf{70.33}  & 58.38  & 48.16  & 45.45 \\
PGD-$l_{\infty}$: 2 & \textbf{32.99}    & 60.75      & 87.75 & 89.35  & 68.78  & 51.99  & 40.69  & 37.83  \\
PGD-$l_{\infty}$: 1 & 32.67    & 61.85      & 89.36 & \textbf{90.18}  & 69.07  & 50.05  & 37.98  & 34.80 \\
PGD-$l_{2}$: 12     & 31.38    & 53.07      & 82.10 & 83.89  & 67.06  & 58.51  & 52.45  & 50.75 \\
PGD-$l_{2}$: 8      & 32.82    & 56.65      & 85.01 & 86.09  & 68.90  & 58.75  & 51.59  & 49.30  \\
PGD-$l_{2}$: 4      & 32.82    & 58.77      & 86.30 & 86.36  & 67.94  & 53.68  & 44.43  & 41.98  \\
FGSM: 8             & 29.53    & 55.46      & 85.10 & 86.65  & 69.01  & 55.64  & 45.92  & 43.42 \\
FGSM: 4             & 32.68    & 59.37      & 87.22 & 87.90  & 66.71  & 51.13  & 41.66  & 38.78 \\
\bottomrule
\end{tabular}
\end{sc}
\end{small}
\end{center}
\vskip -0.1in
\end{table*}

\begin{table*}[h!]
\caption{``Accuracy on correctly classified images''  for different models on saturated Tiny ImageNet test set. It is easily observed AT-CNNs are much more robust to increasing saturation levels on Tiny ImageNet. }
\label{tab:tiny-sat-all}
\vskip -0.15in
\begin{center}
\begin{small}
\begin{sc}
\begin{tabular}{lcccccccc}
\toprule
Saturaion level & 0.25 & 0.5 & 1 & 4 & 8 & 16 & 64 & 1024 \\
\midrule
Standard    & 7.24  & 25.88 & 72.52  & 72.73  & 25.38  & 8.24  & 2.62  & 1.93  \\
Underfit & 7.34  & 25.44  & 69.80 &  60.67  & 18.01  & 6.72  & 3.16  & 2.65                 \\
PGD-$l_{\infty}$: 8 & 11.07  & 29.08  & 67.11 & 74.53  & 49.8  & 40.16  & 35.44  & 33.96   \\
PGD-$l_{\infty}$: 4 & 12.44  & 33.53  & 72.94 & 75.75  & 46.38  & 32.12  & 24.92  & 22.65 \\
PGD-$l_{\infty}$: 2 & 12.09  & 34.85  & 75.77 & 76.15  & 41.35  & 25.20  & 16.93  & 14.52  \\
PGD-$l_{\infty}$: 1 & 11.30  & \textbf{35.03}  & 76.85 & 78.63  & 40.48  & 21.37  & 12.70  & 10.81 \\
PGD-$l_{2}$: 12     & 11.30  & 29.48  & 66.94 & 75.22  & \textbf{52.26 } & \textbf{42.11 } & \textbf{37.20}  & \textbf{35.85} \\
PGD-$l_{2}$: 8      & 12.42  & 32.78  & 71.94 & 75.15  & 47.92  & 35.66  & 29.55  & 27.90  \\
PGD-$l_{2}$: 4      & \textbf{12.63}  & 34.10  & 74.06 & 77.32  & 45.00  & 28.73  & 20.16  & 18.04  \\
FGSM: 8             & 12.59  & 32.66  & 70.55 & \textbf{81.53}  & 41.83  & 17.52  & 7.29  & 5.82 \\
FGSM: 4             & \textbf{12.63 } & 34.10  & \textbf{74.06} & 75.05  & 42.91  & 29.09  & 22.15  & 20.14 \\
\bottomrule
\end{tabular}
\end{sc}
\end{small}
\end{center}
\vskip -0.1in
\end{table*}

\begin{table*}[h!]
\caption{``Accuracy on correctly classified images''  for different models on saturated CIFAR-10 test set. It is easily observed AT-CNNs are much more robust to increasing saturation levels on CIFAR-10.}
\label{tab:cifar-sat-all}
\vskip -0.15in
\begin{center}
\begin{small}
\begin{sc}
\begin{tabular}{lcccccccc}
\toprule
Saturaion level & 0.25 & 0.5 & 1 & 4 & 8 & 16 & 64 & 1024 \\
\midrule
Standard    & 27.36  & 55.95  & \textbf{91.03}  & \textbf{93.12}  & 69.98  & 48.30  & 34.39  & 31.06     \\
Underfit    & 21.43  & 50.28  & 87.71  & 89.89  & 66.09  & 43.35  & 29.10  & 26.13      \\
PGD-$l_{\infty}$: 8 & 26.05  & 46.96  & 80.97  & 89.16  & \textbf{75.46 } & \textbf{69.08}  & 58.98  & \textbf{64.64}   \\
PGD-$l_{\infty}$: 4 & 27.22  & 49.81  & 84.16  & 89.79  & 73.89  & 65.35  & 59.99  & 58.47 \\
PGD-$l_{\infty}$: 2 & \textbf{28.32}  & 53.12  & 86.93  & 91.37  & 74.02  & 62.82  & 55.25  & 52.60  \\
PGD-$l_{\infty}$: 1 & 27.18  & 53.59  & 88.54  & 91.77  & 72.67  & 58.39  & 47.25  & 41.75 \\
PGD-$l_{2}$: 12     & 25.99  & 46.92  & 81.72  & 88.44  & 73.92  & 66.03  & 60.98  & 59.41 \\
PGD-$l_{2}$: 8      & 27.75  & 50.29  & 83.76  & 80.92  & 73.17  & 64.83  & 58.64  & 46.94  \\
PGD-$l_{2}$: 4      & 27.26  & 51.17  & 85.78  & 90.08  & 73.12  & 61.50  & 52.04  & 48.79  \\
FGSM: 8             & 25.50  & 46.11  & 81.72  & 87.67  & 74.22  & 67.12  & \textbf{62.51 } & 61.32  \\
FGSM: 4             & 26.39  & \textbf{58.93}  & 84.30  & 89.02  & 73.47  & 64.43  & 58.80  & 56.82 \\
\bottomrule
\end{tabular}
\end{sc}
\end{small}
\end{center}
\vskip -0.1in
\end{table*}

\begin{table*}[h!]
\caption{``Accuracy on correctly classified images''  for different models on Patch-shuffled Caltech-256 test set. Results indicates that  AT-CNNs are more sensitive to Patch-shuffle operations on Caltech-256. }
\label{tab:patch-cal}
\vskip -0.15in
\begin{center}
\begin{small}
\begin{sc}
\begin{tabular}{lccc}
\toprule
Data set & $2 \times 2$ & $4 \times 4$ & $8 \times 8$t  \\
\midrule
Standard    & \textbf{84.76 } & \textbf{51.50}  & \textbf{10.84} \\
Underfit    & 75.59  & 33.41  & 6.03  \\
PGD-$l_{\infty}$: 8 & 58.13  & 20.14  & 7.70  \\
PGD-$l_{\infty}$: 4 & 68.54  & 26.45  & 8.18  \\
PGD-$l_{\infty}$: 2 & 74.25  & 30.77  & 9.00  \\
PGD-$l_{\infty}$: 1 & 78.11  & 35.03  & 8.42  \\
PGD-$l_{2}$: 12     & 58.25  & 21.03  & 7.85  \\
PGD-$l_{2}$: 8      & 63.36  & 22.19  & 8.48  \\
PGD-$l_{2}$: 4      & 69.65  & 28.21  & 7.72  \\
FGSM: 8             & 64.48  & 22.94  & 8.07 \\
FGSM: 4             & 70.50  & 28.41  & 6.03 \\
\bottomrule
\end{tabular}
\end{sc}
\end{small}
\end{center}
\vskip -0.1in
\end{table*}

\begin{table*}[h!]
\caption{``Accuracy on correctly classified images''  for different models on Patch-shuffled Tiny ImageNet test set. Results indicates that  AT-CNNs are more sensitive to Patch-shuffle operations on Tiny ImageNet.}
\label{tab:path-tiny}
\vskip -0.15in
\begin{center}
\begin{small}
\begin{sc}
\begin{tabular}{lccc}
\toprule
Data set & $2 \times 2$ & $4 \times 4$ & $8 \times 8$t  \\
\midrule
Standard    & \textbf{66.73} & \textbf{24.87}  & 4.48  \\
Underfit    & 59.22  & 23.62  & 4.38  \\
PGD-$l_{\infty}$: 8 & 41.08  & 16.05  & 6.83  \\
PGD-$l_{\infty}$: 4 & 49.54  & 18.23  & 6.30  \\
PGD-$l_{\infty}$: 2 & 55.96  & 19.95  & 5.61  \\
PGD-$l_{\infty}$: 1 & 60.19  & 23.24  & 6.08  \\
PGD-$l_{2}$: 12     & 42.23  & 16.95  & \textbf{7.66}  \\
PGD-$l_{2}$: 8      & 47.67  & 16.28  & 6.50  \\
PGD-$l_{2}$: 4      & 51.94  & 17.79  & 5.89  \\
FGSM: 8             & 57.42  & 20.70  & 4.73 \\
FGSM: 4             & 50.68  & 16.84  & 5.98 \\
\bottomrule
\end{tabular}
\end{sc}
\end{small}
\end{center}
\vskip -0.1in
\end{table*}

\section{Additional Figures}
\begin{figure*}[ht!]
\centering
\begin{tabular}{c}
\includegraphics[width=2\columnwidth]{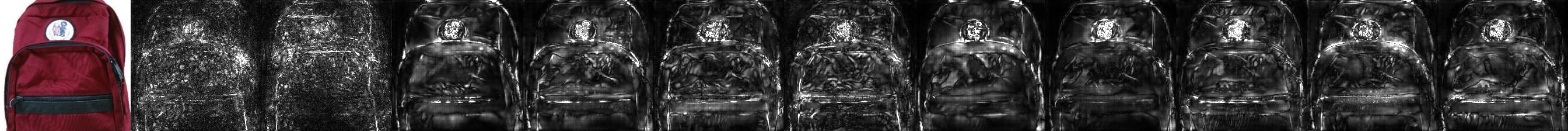}  \\
\includegraphics[width=2\columnwidth]{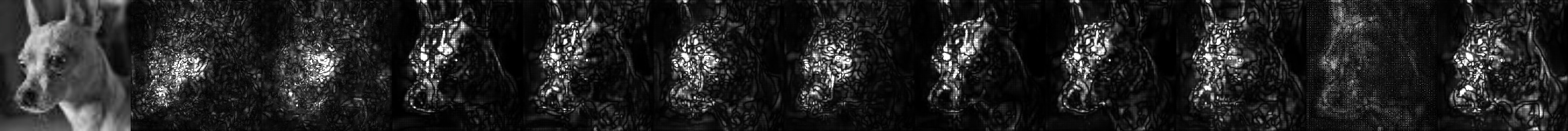}  \\
\includegraphics[width=2\columnwidth]{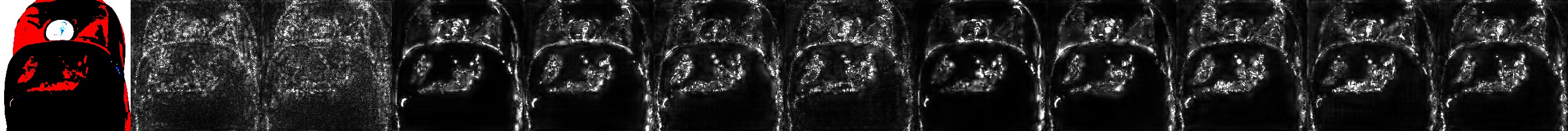}  \\
\includegraphics[width=2\columnwidth]{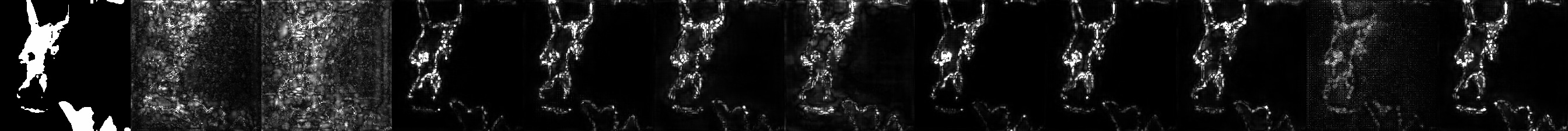}  \\
\includegraphics[width=2\columnwidth]{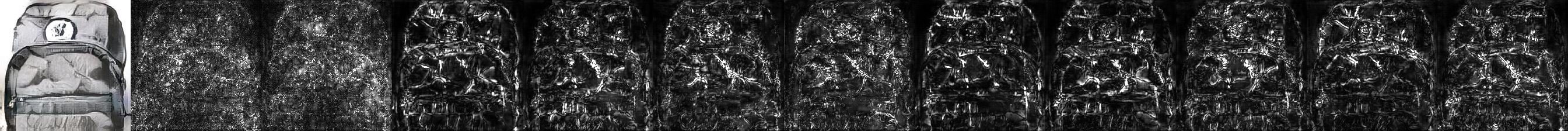}  \\
\includegraphics[width=2\columnwidth]{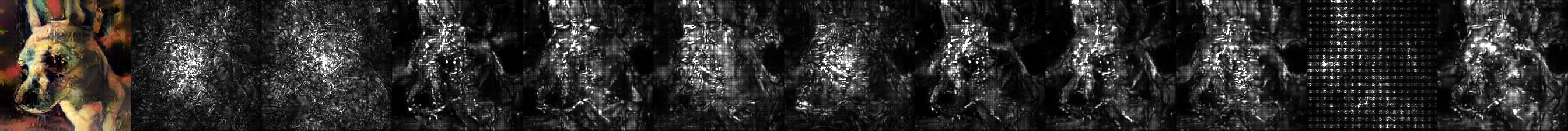} \\

\end{tabular}
\vskip -0.1in
\caption{Visualization of Salience maps generated from \textbf{SmoothGrad} \cite{smilkov2017smoothgrad} for all 11 models. From left to right, Standard CNNs, underfitting CNNs, PGD-inf: 8, 4, 2, 1, PGD-L2: 12, 8, 4 and FGSM: 8, 4.}
\label{fig:all-salience}
\end{figure*}

\begin{figure*}[ht!]
\centering
\begin{tabular}{c}
\includegraphics[width=2\columnwidth]{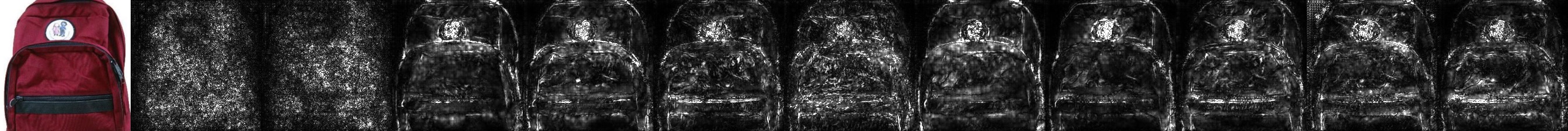}  \\
\includegraphics[width=2\columnwidth]{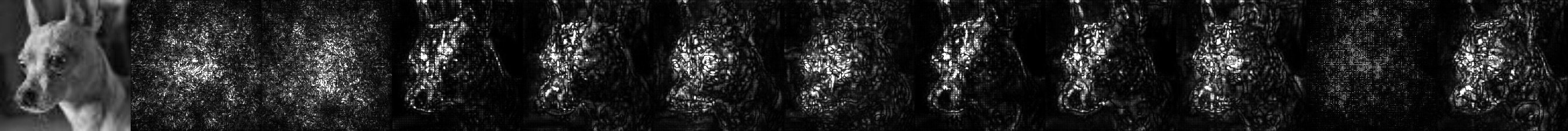}  \\
\includegraphics[width=2\columnwidth]{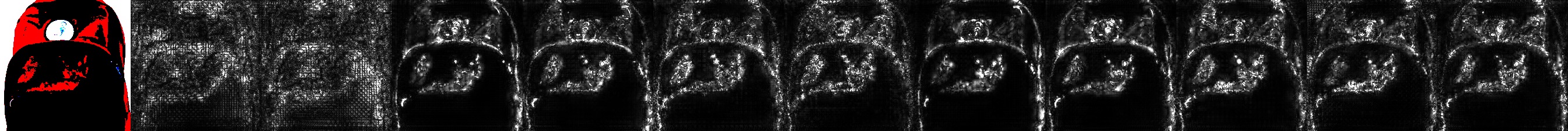}  \\
\includegraphics[width=2\columnwidth]{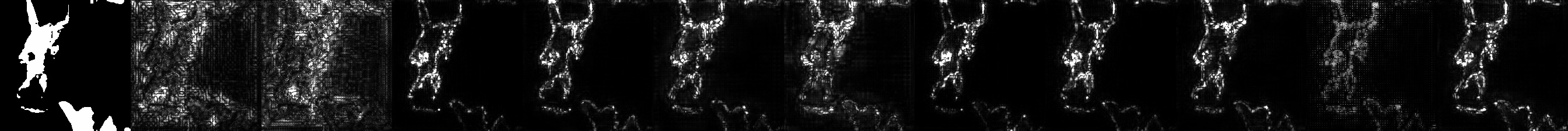}  \\
\includegraphics[width=2\columnwidth]{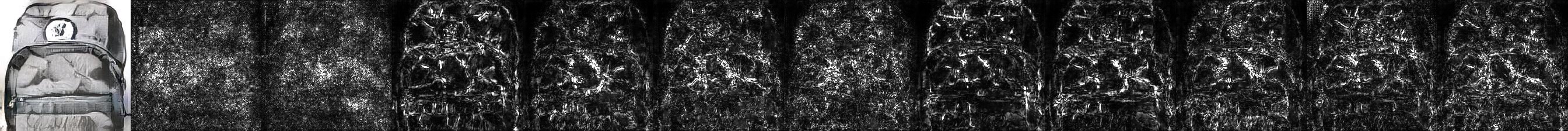}  \\
\includegraphics[width=2\columnwidth]{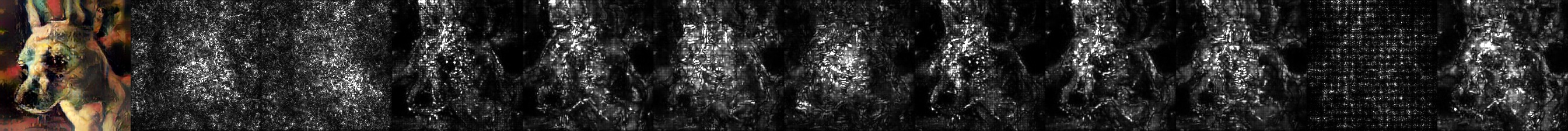} \\

\end{tabular}
\vskip -0.1in
\caption{Visualization of Salience maps generated from \textbf{Grad} for all 11 models. From left to right, Standard CNNs, underfitting CNNs, PGD-inf: 8, 4, 2, 1, PGD-L2: 12, 8, 4 and FGSM: 8, 4. It's easily observed that sensitivity maps generated from \textbf{Grad} are more noisy compared with its smoothed variant \textbf{SmoothGrad}, especially for Standard CNNs and underfitting  CNNs.} 
\label{fig:all-grad}
\end{figure*}

We show additional sensitive maps in Figure~\ref{fig:all-salience}. We also compare the sensitive maps using \textbf{Grad} and \textbf{SmoothGrad} in Figure~\ref{fig:all-grad}.

\bibliography{atcnn}
\bibliographystyle{icml2019}

\end{document}